\pdfoutput=1
\documentclass{article}

\usepackage[dblblindworkshop,final]{neurips_2026}

\usepackage[utf8]{inputenc}
\usepackage[T1]{fontenc}
\usepackage[hidelinks]{hyperref}
\usepackage{url}
\usepackage{booktabs}
\usepackage{amsfonts}
\usepackage{amsmath}
\usepackage{amssymb}
\usepackage{amsthm}
\usepackage{nicefrac}
\usepackage{microtype}
\usepackage{xcolor}
\usepackage{multirow}
\usepackage{graphicx}
\usepackage{tikz}
\usetikzlibrary{arrows.meta,positioning,fit,backgrounds,calc}

\definecolor{agentDesign}{HTML}{E7EEF7}
\definecolor{agentDesignLine}{HTML}{1F4E79}
\definecolor{agentVerify}{HTML}{E5F1EC}
\definecolor{agentVerifyLine}{HTML}{2F6B55}
\definecolor{agentSearch}{HTML}{EFEAF6}
\definecolor{agentSearchLine}{HTML}{4A2C7A}

\theoremstyle{definition}
\newtheorem{definition}{Definition}

\newcommand{\RCS}{\mathrm{RCS}}
\newcommand{\Rhat}{\widehat{\mathrm{RCS}}}

\title{When Can You Trust an Agent's Own Verifier?\\
Stress-Testing Counterfactual Rationale Verification\\
Under Search Pressure and Oracle Access}

\workshoptitle{Who Verifies the Agents? Toward Reliable Agent Development}

\author{%
  Jonghyun Lee \\
  Yonsei University \\
  \texttt{devjh@yonsei.ac.kr} \\
  \And
  Moonsu Kim \\
  Yonsei University \\
  \texttt{moonsu88@yonsei.ac.kr} \\
  \And
  Sunghwan Park \\
  Yonsei University \\
  \texttt{patrick.park@yonsei.ac.kr} \\
}

\begin{document}

\maketitle

\begin{abstract}
Automated agent design increasingly trusts a single score to decide whether a change is an
improvement --- but a score optimized aggressively enough ceases to measure what it was meant to. When
an agent also \emph{explains} its design, that explanation is a second signal we can check
without human labels: negate each stated claim, re-run, and observe whether performance drops.
We ask whether such a verifier stays reliable once an agent optimizes against it. Our contribution is
not a new verifier but a way to \emph{stress-test} one: a pressure ratio $\rho$ measuring how much the
verifier steers the agent's search, and four falsifiable probes covering four failure modes --- giving
no signal, being too weak to matter, being gamed by stating fewer claims, and being queried before the
agent commits. We apply the test to our own verifier, the Rationale Consistency Score (RCS), inside an
LLM agent that designs machine-learning pipelines, with one generator model, six tasks, and three
seeds. First, in our prior configuration the verifier barely affects the search \emph{objective}
($\rho \ll 1$), partly because of an accuracy gate, so its robustness had not been tested; it
still shapes the reported artifact through a final-selection tie-break that $\rho$ does not register.
Second, a capability control shows the agent shortens its claim lists when instructed, yet search does
not, even at $\rho$ more than $100\times$ the prior level; since feedback omits claim
volume and horizons are short, this null result is largely structural rather than evidence of robustness.
Third, when the agent can probe the verifier before committing, it keeps claims that pass while
held-out faithfulness falls --- a decline invisible to in-loop metrics, though partly confounded by
how RCS scores small pipelines. We offer these as preliminary evidence and propose a reporting
standard: audit a verifier with $\rho$ and these probes before trusting it.
\end{abstract}

\section{Introduction}

Verification is the bottleneck in automated agent design. When an outer loop proposes changes to an
agent and an inner loop scores them, the score is the only safeguard separating genuine
improvement from silent regression. In practice that score is usually a scalar task reward, and a
scalar reward optimized aggressively enough ceases to measure the quantity it was intended to
track~\citep{manheim2018categorizing,skalse2022reward,gao2023scaling}.

Agents that design artifacts often also \emph{explain} them: an agent proposing a machine-learning
pipeline states which features matter and why a model class suits the data. That reasoning is usually
discarded, but it has a property the task reward lacks --- it is \emph{checkable against the
environment without human labels}. If the agent claims a feature is important, we remove it, retrain,
and observe whether performance falls: remove-and-retrain evaluation of attribution
methods~\citep{hooker2019benchmark}, applied to natural-language claims rather than saliency maps.

The Rationale Consistency Score (RCS) formalizes this check. It is attractive as a verification
signal for three reasons: it is grounded in the environment rather than in a learned reward model,
it is orthogonal in construction to task accuracy, and it is defined per-claim, so it degrades
gracefully when only part of an explanation is faithful.

For these same reasons, however, RCS is also a target. Once RCS enters an optimization objective,
the agent has an incentive to make claims that survive verification rather than claims that are
informative. The simplest such strategy requires no sophistication: \emph{state fewer claims} --- one
safe claim, not eight, scores well on an average-over-claims metric while communicating almost
nothing. This is specification gaming in its simplest form~\citep{amodei2016concrete,pan2022effects},
undetectable by any evaluation that reports only the verifier's mean output.

This paper asks whether a counterfactual rationale verifier remains reliable when an agent optimizes
against it --- whether it prevents reward hacking and stays faithful as the agent evolves, separately
from whether it improves task performance. Such a signal should be characterized the way a measuring
instrument is --- by its failure modes, operating range, and behavior under adversarial pressure ---
before it is recommended as infrastructure. Our contribution is therefore a \emph{stress test}, not a
verifier: a pressure metric and four falsifiable probes for any counterfactual rationale verifier.

Our findings are preliminary and mixed: in our prior configuration the verifier is too weak to have
been tested (F2); a capability control shows the agent \emph{can} shorten its claim lists when
instructed, yet search never does so even at $\rho \gg 1$ (F3), a null result largely due to short horizons and
feedback that omits claim counts; and the failure that does appear is oracle
exploitation (F4).

\paragraph{Contributions.}
\begin{itemize}
\item \textbf{A stress-test methodology for rationale verifiers}: four failure modes
(Section~\ref{sec:failures}), falsifiable probes for them (Section~\ref{sec:protocol}), and a
capability control for F3, all reusable beyond RCS.
\item The \emph{pressure ratio} $\rho$ (Section~\ref{sec:pressure}), and how an
accuracy-gating rule suppresses it: our prior configuration lies at $\rho \approx 0.007$--$0.089$
(Section~\ref{sec:r-pressure}), while controlled sweeps reach $\rho \approx 41.7$--$63.7$
(Section~\ref{sec:r-dose}).
\item A capability control showing that the agent shortens its claim lists when instructed
(Section~\ref{sec:r-capability}) while search does not, even at $\rho \gg 1$ and under UCB1 at a
doubled base horizon (Section~\ref{sec:r-dose}) --- a discovery gap, not a demonstrated reward exploit.
\item Evidence that probe access produces the selection signature of Section~\ref{sec:oracle} on four
of six datasets at sign level, with held-out fidelity declines that the estimator may partly explain
(Section~\ref{sec:r-oracle}).
\end{itemize}

\section{Related Work}
\label{sec:related}

\paragraph{Reward hacking and Goodhart's law.} That a proxy optimized aggressively enough ceases to track its
target recurs in reward misspecification~\citep{manheim2018categorizing,skalse2022reward} and
reward-model overoptimization~\citep{gao2023scaling}, and claim minimization is an instance of
specification gaming~\citep{amodei2016concrete,pan2022effects}. The closest structural
precedent is a learned judge or reward model folded into an objective: LLM judges carry exploitable
biases~\citep{zheng2023judging,wang2024fair}, and policies over-optimize learned
rewards~\citep{gao2023scaling,coste2024reward}. RCS replaces the learned judge with environment
interventions; we audit it, measuring how much pressure the verifier exerts before asking whether it is gamed.

\paragraph{Rationale faithfulness.} Whether stated reasoning reflects the underlying computation has
motivated a growing literature: chain-of-thought can be systematically
unfaithful~\citep{turpin2023language,lanham2023measuring}, and faithfulness is distinct from
plausibility~\citep{jacovi2020towards}. RCS operationalizes it through counterfactual
interventions~\citep{chuang2024faithlm}, extending remove-and-retrain evaluation from saliency
maps~\citep{hooker2019benchmark} and feature attribution~\citep{ribeiro2016why,lundberg2017unified}
to natural-language claims. Such removal-based metrics are themselves gameable under
optimization~\citep{hsia2024goodhart}; RCS inherits that exposure, which is why we audit it under search
pressure rather than assuming robustness.

\paragraph{LLM-driven pipeline design.} Automated pipeline construction has moved from search- and ensemble-based
systems~\citep{olson2016tpot,erickson2020autogluon} to LLM optimizers~\citep{yang2023llm} and proposers
that suggest features in natural language~\citep{hollmann2023caafe}. Our agent adds a verifier over its stated
reasons --- precisely what creates the gaming surface we study.

\section{Setting}
\label{sec:setting}

\paragraph{The design task.} We instantiate automated agent design in a constrained but fully
executable environment: the agent designs supervised learning pipelines $P = (S, M, H)$ --- a feature
subset $S \subseteq F$, a model class $M$, and hyperparameters $H$ --- with task performance
$Q(P) \in [0,1]$ estimated as $\hat{Q}(P)$ on validation data. The setting is narrow relative to
open-ended agentic tasks (Section~\ref{sec:limitations}), but every claim the agent makes has a
mechanically constructible counterfactual, which makes the verifier study feasible.

\paragraph{The agent.} The agent is a large language model inside an optimization loop. In the
default proposer --- the one used for the main configuration, the pressure sweeps, and the capability
control --- each iteration is a single call that returns a feature set, a model, hyperparameters, and
a natural-language rationale $E_P$. The tool-using proposer, which may call \texttt{list\_features},
optionally \texttt{probe\_feature\_drop}, and then \texttt{submit\_pipeline}, appears only in the
probe-access probe (Section~\ref{sec:oracle}), where both arms use it. Proposals are executed, scored, and used to update Thompson sampling posteriors
over pipeline-type arms~\citep{agrawal2012analysis,kaufmann2012thompson}, whose recommendations are
fed back into the next prompt.

\paragraph{Claims.} The rationale $E_P$ is parsed into a set of claims
$E_P = \{C_1,\dots,C_k\}$ of three kinds: feature-importance claims, model-choice claims, and
hyperparameter claims. Parsing is deterministic and does not depend on the verification machinery,
a property we rely on in Section~\ref{sec:instrumentation}.

\begin{figure*}[t]
\centering
\resizebox{\textwidth}{!}{%
\begin{tikzpicture}[
  font=\fontsize{11.5}{13.8}\selectfont,
  bx/.style 2 args={draw=#1, line width=0.9pt, fill=white, rounded corners=1.5pt,
                    align=center, minimum height=1.75cm, minimum width=#2, inner sep=4pt},
  ar/.style={-{Latex[length=2.6mm,width=1.8mm]}, line width=0.9pt, draw=black!82},
  arO/.style={ar, dash pattern=on 0.1pt off 2.8pt, line cap=round,
              draw=agentDesignLine, line width=1.7pt},
  arF/.style={ar, dash pattern=on 7pt off 3.5pt,
              draw=agentSearchLine, line width=1.1pt},
  msg/.style={font=\fontsize{11.5}{13.5}\selectfont, fill=white, inner sep=2.5pt},
  msgO/.style={msg, text=agentDesignLine},
  msgF/.style={msg, text=agentSearchLine},
  tag/.style={font=\fontsize{11}{12.5}\selectfont\bfseries, fill=black!6,
              draw=black!45, rounded corners=1pt, inner sep=2pt},
  role/.style={font=\fontsize{13.5}{15.5}\selectfont\bfseries, align=center},
  lgd/.style={anchor=west, font=\fontsize{11.5}{13.5}\selectfont},
  lane/.style={draw=#1, line width=0.9pt, rounded corners=2pt}
]

\fill[agentDesign] (1.50,6.30) rectangle (13.95,8.60);
\draw[lane=agentDesignLine!45] (1.50,6.30) rectangle (13.95,8.60);
\fill[agentVerify] (1.50,3.20) rectangle (13.95,5.50);
\draw[lane=agentVerifyLine!45] (1.50,3.20) rectangle (13.95,5.50);
\fill[agentSearch] (1.50,0.10) rectangle (13.95,2.40);
\draw[lane=agentSearchLine!45] (1.50,0.10) rectangle (13.95,2.40);

\node[role,text=agentDesignLine] at (0.68,7.45) {Design\\Agent};
\node[role,text=agentVerifyLine] at (0.68,4.35) {Verifier\\Agent};
\node[role,text=agentSearchLine] at (0.68,1.25) {Search\\Agent};


\node[bx={agentDesignLine}{3.2cm}] (llm) at (3.65,7.45) {\textbf{LLM proposer}};
\node[bx={agentDesignLine}{4.2cm}] (tools) at (8.35,7.45)
  {\textbf{Tool interface}\\[2pt]
   \textcolor{black!25}{\rule{3.6cm}{0.4pt}}\\[2pt]
   {\fontsize{10.5}{12.5}\selectfont\ttfamily list\_features}\\[1pt]
   {\fontsize{10.5}{12.5}\selectfont\ttfamily probe\_feature\_drop}};
\node[bx={agentDesignLine}{2.9cm}] (submit) at (12.45,7.45)
  {\textbf{Submit}\\[3pt] $(S,M,H)+E_P$};
\draw[ar] (llm) -- (tools);
\draw[ar] (tools) -- (submit);

\node[bx={agentVerifyLine}{2.8cm}] (parse) at (12.45,4.35) {\textbf{Claim parser}};
\node[tag] at ($(parse.north east)+(-0.44,-0.34)$) {F3};
\node[bx={agentVerifyLine}{3.2cm}] (cf) at (8.35,4.35)
  {\textbf{Counterfactual}\\\textbf{constructor}};
\node[bx={agentVerifyLine}{2.4cm}] (exec) at (5.15,4.35) {\textbf{Executor}\\\textbf{(retrain)}};
\node[bx={agentVerifyLine}{1.9cm}] (rcs) at (2.60,4.35) {$\Rhat$};
\node[tag] at ($(rcs.north east)+(-0.42,-0.34)$) {F1};
\draw[ar] (parse) -- (cf);
\draw[ar] (cf) -- (exec);
\draw[ar] (exec) -- (rcs);

\node[bx={agentSearchLine}{4.2cm}] (obj) at (5.15,1.25) {$R=\hat{Q}+\lambda_t\,\Rhat$};
\node[tag] at ($(obj.north east)+(-0.44,-0.34)$) {F2};
\node[bx={agentSearchLine}{2.7cm}] (ts) at (12.45,1.25)
  {\textbf{Thompson}\\\textbf{Sampling}};
\draw[ar] (obj) -- node[msg,pos=0.5] {reward} (ts);

\draw[ar] (submit.south) -- node[msg,pos=0.5] {artifact} (parse.north);
\draw[arO] (tools.south) -- node[msgO,pos=0.5] {F4 oracle} (8.35,5.225);
\draw[ar] (exec.south) -- node[msg,pos=0.5] {$\hat{Q}$} (obj.north);
\draw[ar] (rcs.south) -- node[msg,pos=0.30] {$\Rhat$} (2.60,1.25) -- (obj.west);

\draw[arF] (ts.east) -- (14.45,1.25) -- (14.45,9.05)
  -- node[msgF,pos=0.45] {next iteration} (3.65,9.05) -- (llm.north);

\draw[line width=1.0pt,draw=black!82] (1.80,-0.58) -- (3.05,-0.58);
\node[lgd] at (3.15,-0.58) {within-agent / scored messages};
\draw[line width=1.8pt,dash pattern=on 0.1pt off 2.8pt,line cap=round,
      draw=agentDesignLine] (1.80,-1.06) -- (3.05,-1.06);
\node[lgd] at (3.15,-1.06) {F4 oracle channel};
\draw[line width=1.2pt,dash pattern=on 7pt off 3.5pt,draw=agentSearchLine]
  (8.30,-1.06) -- (9.55,-1.06);
\node[lgd] at (9.65,-1.06) {search feedback};
\end{tikzpicture}}
\caption{Multi-agent architecture of the stress-test loop. The \textbf{Design Agent} proposes a
pipeline and rationale in a single call, and through the tool interface only in the P2 arms
(Section~\ref{sec:oracle});
the \textbf{Verifier Agent} parses claims, builds counterfactuals, and returns
$\Rhat$; the \textbf{Search Agent} forms
$R=\hat{Q}+\lambda_t\Rhat$ and updates Thompson Sampling.
Cross-agent edges carry the artifact, oracle probes
(F4, oracle arm only), task/verifier scores, and the next-iteration
recommendation. Tags F1--F4 mark where each protocol probe attaches.}
\label{fig:architecture}
\end{figure*}
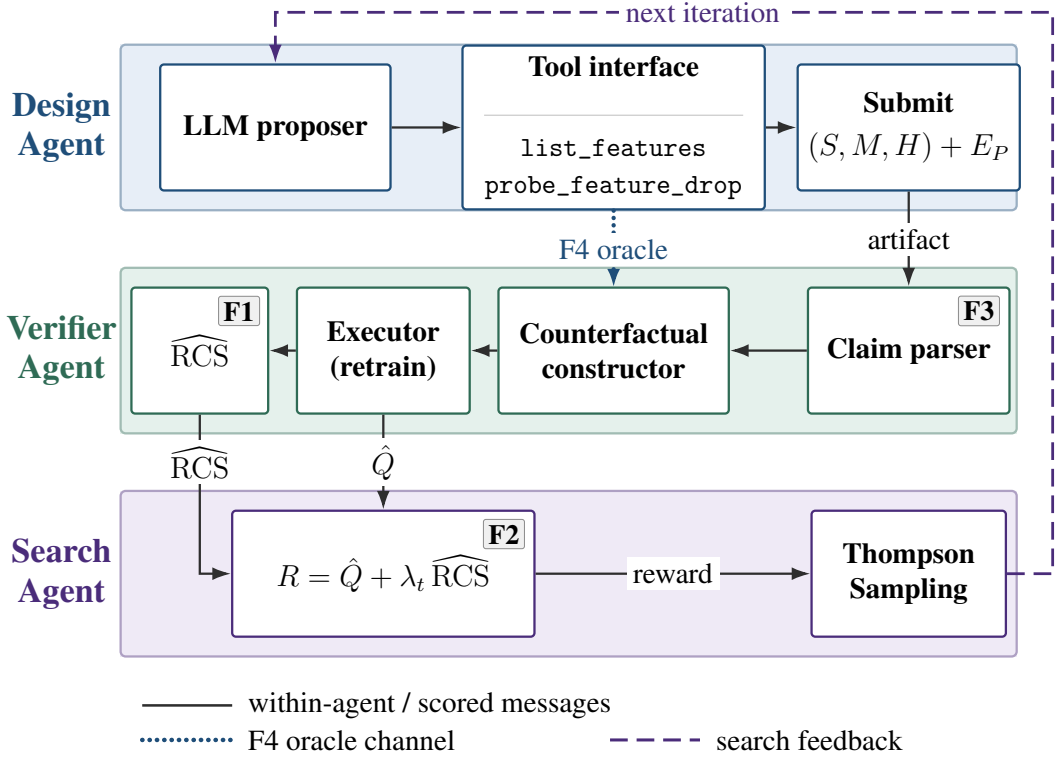

\section{RCS as a Verifier}
\label{sec:rcs}

\paragraph{Provenance.} RCS as defined in Equation~\ref{eq:rcs} is introduced here; we are not auditing
a third-party standard. The system we instrument is our own research agent, and \textbf{``deployed''
throughout means our prior configuration} of that agent --- including the accuracy-gating rule of
Section~\ref{sec:pressure} --- so the F2 measurement is a self-audit and a motivating case, not a survey finding.

\begin{definition}[Rationale Consistency Score]
For each claim $C \in E_P$, let $P^{\neg C}$ denote a counterfactual pipeline that contradicts $C$
while holding other design choices fixed. Then
\begin{equation}
\RCS(P) \;=\; \frac{1}{|E_P|} \sum_{C \in E_P} \max\!\left(0,\; Q(P) - Q(P^{\neg C})\right).
\label{eq:rcs}
\end{equation}
\end{definition}

Counterfactuals are constructed per claim type: a feature-importance claim is negated by removing
the feature and retraining; a model-choice claim by substituting an alternative model class; a
hyperparameter claim by perturbing that hyperparameter. When three or more feature-importance claims
are stated, the estimator groups them into one joint intervention (Appendix~\ref{app:rcs}).

\paragraph{What RCS verifies, and what it does not.} RCS asks whether the agent's stated reasons
are load-bearing for the artifact it produced. It does not ask whether the artifact performs well, and it
does not certify that a change to the agent improved capability. It is therefore a
\emph{heterogeneous signal} complementing outcome-based verification rather than replacing it.

\paragraph{Estimator variants.} Replacing $Q$ by the empirical $\hat{Q}$ gives the base estimator
\begin{equation}
\Rhat_{\text{base}}(P) \;=\; \frac{1}{|E_P|}\sum_{C \in E_P}
  \max\!\left(0,\; \hat{Q}(P) - \hat{Q}(P^{\neg C})\right),
\label{eq:rcs-base}
\end{equation}
which is sensitive to validation noise. The deployed system uses a practical variant with a small
modality-dependent noise threshold, a relative-drop boost, and an amplification threshold
(Appendix~\ref{app:rcs}); Equations~\ref{eq:rcs}--\ref{eq:rcs-base} idealize it; all results
concern this deployed variant, abbreviated $\Rhat$.

\section{How a Rationale Verifier Fails}
\label{sec:failures}

We distinguish four failure modes. The first two concern the verifier as an instrument; the last two
concern the agent's response to it. They follow from the structure of the setting rather than from
RCS's internals (though tested only on RCS): F2 applies to any auxiliary term in a composite objective, F3 to any per-claim average
whose denominator the agent controls, and F4 to any verifier exposed as a callable tool.

\paragraph{F1: Silence.} The verifier returns no usable signal. Every counterfactual drop falls
below the noise threshold, or no claims could be parsed, and the reported score is a floor value
rather than a measurement. We treat the \emph{floor rate} as a primary reported quantity.

\paragraph{F2: Untested.} The verifier appears in the objective but contributes so little relative
to task-reward variation that the agent's search is effectively unchanged. Observing no gaming under
these conditions is uninformative about robustness: the incentive was never applied.

\paragraph{F3: Claim minimization.} Suppose $k_c$ of the $|E_P|$ claims are counterfactually
impactful. Taking expectations in Equation~\ref{eq:rcs} then gives
\begin{equation}
\mathbb{E}\!\left[\RCS(P)\right] \;=\; \frac{k_c}{|E_P|}\,\bar{\Delta}_c ,
\label{eq:minimization}
\end{equation}
with $\bar{\Delta}_c$ the mean drop over impactful claims. The score depends on the \emph{ratio} of
impactful to stated claims, so deleting one inert claim multiplies it by $|E_P|/(|E_P|-1)$ while
leaving the artifact untouched. This holds for $\RCS$; the deployed $\Rhat$ groups feature
claims and averages over executed interventions (Appendix~\ref{app:rcs}), so deleting a stated claim
can leave $\Rhat$ unchanged. We test whether search discovers shorter claim lists under pressure (P1) and
whether the agent produces them when instructed (P4).

\paragraph{F4: Oracle exploitation.} If the agent can query the verifier before committing, it can
filter its claims to those that pass. The resulting rationale is \emph{verified} but selected.
Distinguishing genuine improvement from selection requires observing claim volume alongside pass rate.

\section{A Stress-Test Protocol}
\label{sec:protocol}

Each probe states a treatment, a control, and a falsifying observation. The mapping is not
one-to-one: F2 is read directly off $\rho$, P1 and P4 both target F3, P2 targets F4, and P3 targets F1.

\subsection{Auditing Verifier Pressure}
\label{sec:pressure}

Let the search objective be $R(P) = \hat{Q}(P) + \lambda \cdot \Rhat(P)$. We define the
\textbf{pressure ratio} over a run of $T$ iterations:
\begin{equation}
\rho \;=\; \frac{\frac{1}{T}\sum_{t=1}^{T} \lambda_t \cdot \Rhat(P_t)}
                {\mathrm{sd}\!\left(\{\hat{Q}(P_t)\}_{t=1}^{T}\right)},
\label{eq:pressure}
\end{equation}
where $\lambda_t$ is the weight actually applied at iteration $t$. When $\rho \ll 1$ the verifier has
negligible influence on search. We report $\rho$ for every run.

\paragraph{Three claims of decreasing generality.} (i) \emph{Applied pressure must be measured}: in any
composite objective, whether an auxiliary term steers the search is not readable from its nominal
weight. This is our primary methodological claim. (ii) \emph{$\rho$ is one operationalization of it.}
Its denominator is a design choice that Section~\ref{sec:limitations} shows can inflate, and its
numerator is a time average, so pressure concentrated in a few iterations is indistinguishable from
uniformly small pressure; the selection flip rate we report beside it (Table~\ref{tab:flip}) is exposed
to neither. (iii) \emph{The $\rho \approx 0.007$--$0.089$ comes from a single self-audit}, not a
survey, and partly reflects the gate below; ungated at $\lambda_{\mathrm{nom}}=0.12$, the sweep level
nearest the global nominal $0.1$, it is $0.04$--$0.28$ (Table~\ref{tab:dose}).

\paragraph{Why nominal weights mislead.} The system first scales the nominal weight, then gates it on
validation accuracy (Appendix~\ref{app:rcs}). On Credit-G, $\bar{\lambda}_t=0.030$ versus
$s_d\lambda_{\mathrm{nom}}=0.088$ (global nominal $0.1$), a factor of $0.34$: a plausible gate, not
chosen to lower $\rho$, silently reduces the applied weight. Controlled sweeps disable it but retain
scaling: $R(P)=\hat{Q}(P)+s_d\lambda_{\mathrm{nom}}\Rhat(P)$, where $s_d\in[0.78,0.98]$.

\subsection{P1: Dose--Response over Verifier Pressure}
\label{sec:doseresponse}

\textbf{Question.} Does increasing verifier pressure cause claim minimization (F3)?
\textbf{Design.} Sweep nominal $\lambda_{\mathrm{nom}}\in\{0,0.12,1,5,20\}$ with all else fixed, three
seeds per arm; the $\lambda_{\mathrm{nom}}=0$ arm still computes $\Rhat$ and only omits it from the
reward. \textbf{Falsification.} Declining claim counts with $\lambda_{\mathrm{nom}}$ is evidence of
gaming; stable counts at $\rho \gg 1$ support robustness.

\subsection{P2: Access to the Agent's Own Verifier}
\label{sec:oracle}

\textbf{Question.} When an agent can query the verifier before committing, does its rationale become
more faithful or merely better selected (F4)?
\textbf{Design.} Both arms use the tool-using proposer and may call \texttt{list\_features}; only the
treatment may call \texttt{probe\_feature\_drop}, capped at four calls per iteration.
\textbf{Interpretation, stated before the results.} In Equation~\ref{eq:selection-signature},
$\Delta\hat v>0$ with $\Delta\bar N\approx0$ suggests better reasoning; the displayed conjunction
is our diagnostic selection signature.

\subsection{P3: Reliability and Disclosure of Failure}
\label{sec:p3}

\textbf{Question.} How often does the verifier return no signal (F1), and how stable is it across
seeds? \textbf{Design.} Repeat the main configuration over three seeds and report bootstrap CIs of
held-out RCS and task accuracy together with the floor rate. Held-out RCS is measured post-hoc for all
arms, including those whose objective excludes RCS.

\subsection{P4: Capability Control for Claim Minimization}
\label{sec:p4}

\textbf{Question.} Can the agent reduce claims when the minimization incentive is made explicit?
\textbf{Design.} At fixed nominal $\lambda_{\mathrm{nom}}=5$, three seeds per policy, compare a
\emph{default} prompt with a \emph{minimize} prompt that both discloses the averaging scoring rule and
instructs the agent to state fewer claims while preserving accuracy. Because this arm alters the agent's
information and its instruction together, it bounds capability rather than isolating it: a reduction
shows minimization is \emph{reachable}, not that instruction alone would suffice. If minimize reduces
claims while search under P1 does not, F3's absence under search reflects a discovery gap rather than an
inability.

\section{Instrumentation}
\label{sec:instrumentation}

For arm $a$ with $T_a$ recorded iterations, let $n_{a,t}=|E_{P_{a,t}}|$ be parsed claims,
$m_{a,t}$ tested claims, and $z_{a,t}$ tested claims with positive counterfactual drop. We report
\begin{equation}
\bar N_a=\frac{1}{T_a}\sum_{t=1}^{T_a}n_{a,t},\qquad
\hat v_a=\frac{\sum_t z_{a,t}}{\sum_t m_{a,t}},\qquad
\Delta\bar N<0\ \wedge\ \Delta\hat v>0\ \Longrightarrow\ \text{selection signature},
\label{eq:selection-signature}
\end{equation}
where $\Delta x=x_{\mathrm{treatment}}-x_{\mathrm{control}}$; $\hat v_a$ is N/A if its denominator is
zero. We compute $n_{a,t}$ independently of verification and also log applied $\lambda_t$, floor vs.\
measurement, tool/probe calls, and submission path. Feedback exposes task and verifier scores but not
$n_{a,t}$, so reward has no direct path to claim volume. Released code, prompts, parser, and per-run
logs make $\rho$ and every probe recomputable.

\section{Experimental Setup}
\label{sec:setup}

We run the protocol on six supervised design tasks spanning four modalities (Table~\ref{tab:datasets}).
Each dataset is split $64/16/20$ into train, validation, and test with preprocessing fit on the
training partition only. Search uses validation accuracy; all figures labeled \emph{held-out} are
computed once on the test partition after selection. The generator is a single commercial
instruction-tuned model, Solar Pro~3 of Upstage's SOLAR lineage~\citep{kim2024solar}, accessed through
its hosted API with default decoding parameters and fixed across all arms, so behavioral differences
are attributable to the objective and tool access rather than the model. Each run executes 16--19
iterations depending on dataset budget. All confidence intervals are percentile bootstrap intervals over per-seed run means
($10{,}000$ resamples). All probes use three seeds ($42, 43, 44$). Across the main configuration, pressure sweep (P1),
probe-access probe (P2), and capability control (P4) this yields 180 runs; one run of the main
configuration produces both the RCS-included and RCS-excluded arms. The UCB1 sweep in
Section~\ref{sec:r-dose} adds a further 90.

\paragraph{Identical-configuration replicate.} Because every run resamples the generator, bootstrap
intervals over per-seed means miss run-to-run variation at a \emph{fixed} seed. Two arms of our design
are the same configuration executed twice, which bounds that component
(Table~\ref{tab:replicate}, Appendix~\ref{app:tables}): claim volume differs by up to $11.4\%$, held-out
RCS by $\times0.61$--$\times2.71$, test accuracy by up to $0.048$, and verified-claim rate --- the one
component the table does not list --- by up to $0.06$. We read every effect below against this band.

\begin{table}[t]
\centering
\caption{Design tasks and modalities; sources and feature construction are in Appendix~\ref{app:data}.
$n$ counts the rows actually ingested, which for four tasks exceeds the canonical size by two.}
\label{tab:datasets}
\small
\begin{tabular}{llrrl}
\toprule
Dataset & Modality & $n$ & $d$ & Notes \\
\midrule
OpenML Credit-G & tabular & 1002 & 20 & credit scoring \\
UCI Breast Cancer & tabular & 288 & 9 & binary recurrence \\
Madelon & tabular & 2002 & 500 & synthetic, high-$d$ \\
Ecommerce Text & text & 50427 & 1200 & TF-IDF features \\
Butterfly Image & image & 6499 & 26 & 75-class species ID \\
Electricity & time-series & 45312 & 23 & ELEC2, up/down \\
\bottomrule
\end{tabular}
\end{table}

\paragraph{Held-out RCS reporting.} Internal logs contain clipped summary fields
(Appendix~\ref{app:rcs}); all tables and figures report the raw, unclipped post-hoc estimator output
(\textbf{uncapped}), though the estimator's per-drop rules still apply (Section~\ref{sec:r-oracle}).
Held-out RCS enters neither the objective nor winner selection, so it cannot leak test information into search.

\section{Results}
\label{sec:results}

\subsection{Deployment pressure is low (F2)}
\label{sec:r-pressure}

Table~\ref{tab:main} reports the main configuration with dynamic $\lambda$. Measured pressure ranges
from $\rho = 0.007$ (Ecommerce) to $\rho = 0.089$ (Madelon): the verifier moved the objective by
less than nine percent of the accuracy spread it was weighed against on every dataset. This is F2, and
it is not an averaging artifact: no individual run among the 18 exceeds $\rho = 0.10$.

\begin{table}[t]
\centering
\caption{Main configuration with dynamic $\lambda$, three seeds per dataset. Brackets: bootstrap 95\%
CIs. Held-out RCS is an uncapped raw measure.}
\label{tab:main}
\footnotesize
\setlength{\tabcolsep}{4pt}
\renewcommand{\arraystretch}{1.15}
\begin{tabular}{llcccc}
\toprule
Dataset & Objective & Test acc.\ & Held-out RCS & Claims/iter & $\rho$ \\
\midrule
 \multirow{2}{*}{Credit-G} & RCS included & 0.723 [0.682, 0.751] & 0.037 [0.010, 0.062] & 11.6 & 0.012 \\
  & RCS excluded & 0.697 [0.682, 0.706] & 0.016 [0.002, 0.029] & 11.3 & 0 \\
\addlinespace[3pt]
 \multirow{2}{*}{Breast} & RCS included & 0.724 [0.690, 0.741] & 0.011 [0.004, 0.015] & 9.3 & 0.010 \\
  & RCS excluded & 0.718 [0.707, 0.741] & 0.014 [0.010, 0.020] & 10.1 & 0 \\
\addlinespace[3pt]
 \multirow{2}{*}{Madelon} & RCS included & 0.741 [0.713, 0.771] & 0.109 [0.005, 0.185] & 20.4 & 0.089 \\
  & RCS excluded & 0.759 [0.738, 0.771] & 0.146 [0.101, 0.173] & 17.0 & 0 \\
\addlinespace[3pt]
 \multirow{2}{*}{Ecommerce} & RCS included & 0.675 [0.648, 0.717] & 0.056 [0.024, 0.076] & 9.7 & 0.007 \\
  & RCS excluded & 0.794 [0.754, 0.825] & 0.163 [0.084, 0.303] & 10.8 & 0 \\
\addlinespace[3pt]
 \multirow{2}{*}{Butterfly} & RCS included & 0.189 [0.178, 0.198] & 0.087 [0.053, 0.121] & 14.8 & 0.010 \\
  & RCS excluded & 0.175 [0.134, 0.201] & 0.043 [0.009, 0.067] & 11.7 & 0 \\
\addlinespace[3pt]
 \multirow{2}{*}{Electricity} & RCS included & 0.763 [0.761, 0.765] & 0.009 [0.003, 0.014] & 8.4 & 0.028 \\
  & RCS excluded & 0.757 [0.747, 0.765] & 0.004 [0.003, 0.004] & 8.6 & 0 \\
\bottomrule
\end{tabular}
\end{table}

Including the verifier in the objective moves held-out accuracy and rationale fidelity by no more
than the seed-to-seed spread on most datasets; Ecommerce is the exception in both directions: the
RCS-included arm loses held-out accuracy ($0.675$ vs.\ $0.794$) and held-out fidelity ($0.056$ vs.\
$0.163$), each with non-overlapping CIs (Table~\ref{tab:main}) --- there, placing the verifier in the
objective already degrades what it is meant to protect. Though seen on one dataset, with an accuracy
gap ($0.119$) $2.5\times$ the replicate gap ($0.048$), it is the most direct evidence here that
an in-objective verifier can backfire. On Credit-G the point estimates favor the RCS-included arm on both held-out accuracy
($0.723$ vs.\ $0.697$) and held-out RCS ($0.037$ vs.\ $0.016$), though the CIs overlap; either way, the
excluded arm must be measured post-hoc rather than scored zero. The floor rate (F1) ranges from $0\%$
(Madelon) to $9.3\%$ (Breast Cancer); silence is not the limiting problem --- through the objective
term, the verifier produces measurements that do not steer search. One influence channel lies outside
$\rho$: the deployed configuration also consults RCS as a final-selection tie-break
(Appendix~\ref{app:rcs}). Re-selecting each run's reported pipeline without the
RCS term flips it in $5$ of $18$ runs, but every flip is among accuracy-tied candidates
($0.000$ conceded; Table~\ref{tab:flip}): at deployment the verifier breaks ties rather than trading
accuracy for fidelity.

\subsection{Dose--response: no claim minimization discovered (F3)}
\label{sec:r-dose}

\begin{figure}[t]
\centering
\includegraphics[width=0.92\linewidth]{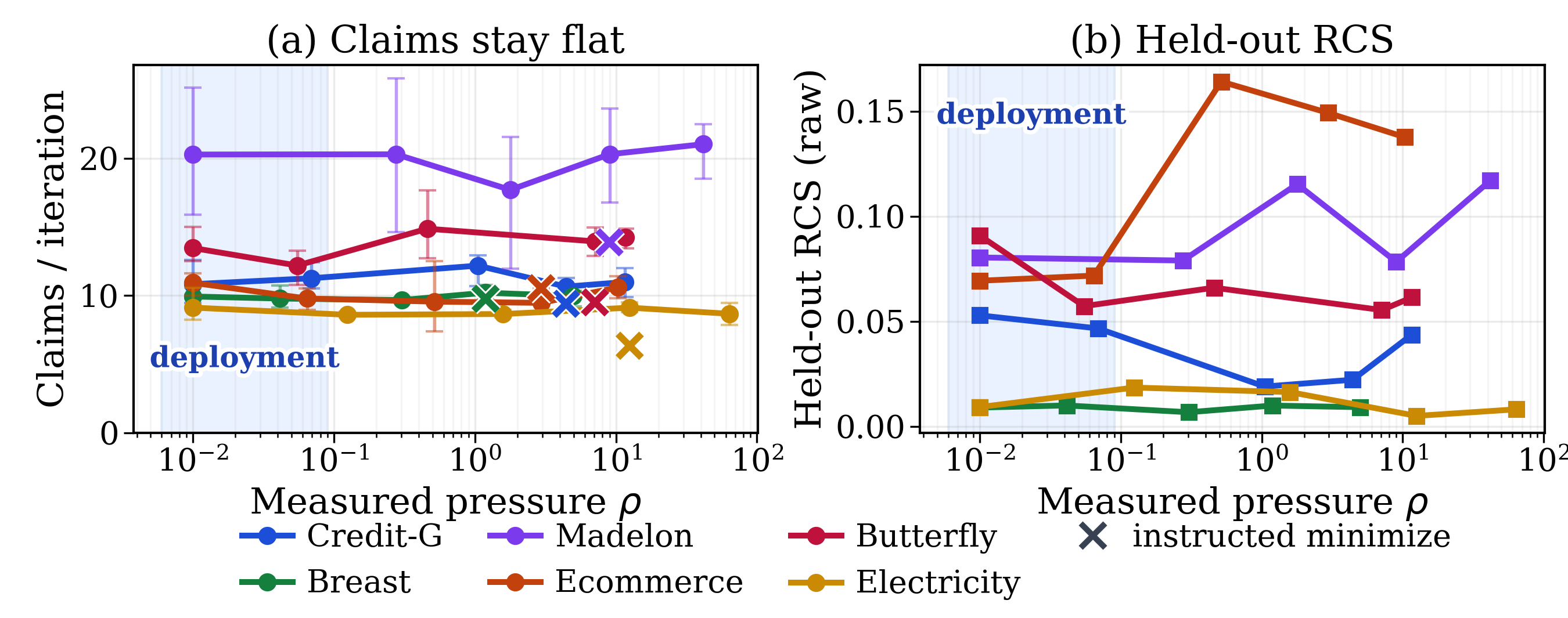}
\caption{Dose--response across six datasets (three seeds each); shading marks deployment. (a) Over
$>2$ orders above it, claims show no significant downward trend (Spearman $\in[-0.12,+0.38]$, all $p>0.15$);
crosses mark lower levels reached under instructed minimization on four responsive datasets.
(b) Uncapped held-out RCS shows no consistent increase with $\rho$.}
\label{fig:dose}
\end{figure}

Figure~\ref{fig:dose} sweeps nominal static $\lambda$ over two orders of magnitude, reaching
$\rho\approx41.7$ on Madelon and $63.7$ on Electricity at $\lambda_{\mathrm{nom}}=20$ --- $2.7$ and
$2.9$ orders above the highest dataset-mean deployed value ($\rho=0.089$; $3.8$ and $4.0$ above the
lowest). Electricity's value additionally
reflects its low accuracy variance inflating $\rho$'s denominator (Section~\ref{sec:limitations}). The
sweep moved decisions and not only the ratio: the reported pipeline flips relative to accuracy-only
selection in $33\%$ of runs at $\lambda_{\mathrm{nom}}=0.12$, $78\%$ at $5$, and $72\%$ at $20$, where a
flip concedes up to $0.19$ validation accuracy, while at $\lambda_{\mathrm{nom}}=0$ it never flips
(Table~\ref{tab:flip}).

\textbf{F3 is not observed under search.} Claim volume shows no statistically significant downward
trend and is nowhere monotonically decreasing in $\lambda_{\mathrm{nom}}$ (Credit-G $10.6$--$12.2$;
Breast Cancer $9.7$--$10.2$; Madelon $17.7$--$21.0$, declining at $\lambda_{\mathrm{nom}}=1$ and
recovering at $20$). A per-dataset Spearman test ($n = 15$) finds no significant negative trend
(correlations $-0.12$ to $+0.38$, all $p > 0.15$; none survives Holm--Bonferroni), and rationale length
is likewise stable (e.g.\ Credit-G $282$--$333$ characters). Altering the search rule and budget does
not change this: replacing Thompson sampling with a deterministic UCB1 bandit~\citep{auer2002finite} and
doubling the base cap to 30 iterations (realized $18$--$38$) leaves claim volume unchanged, peak pressure still
$\rho \approx 39$ (Madelon $38.7$, Electricity $38.4$), and every per-dataset Spearman correlation in $[-0.17, +0.33]$ (all $p > 0.23$;
Table~\ref{tab:dose_ucb}, Appendix~\ref{app:tables}).

The per-dataset tests ($n = 15$) detect only large monotone trends ($|r_s| \approx 0.52$ at
$\alpha = 0.05$, $\approx 0.65$ under Holm). The null is informative not because these tests found
nothing, but because instructed minimization reaches a level below the mean claim volume at every
pressure level on the four datasets where it reduced claims ($18$--$35\%$ below default,
Section~\ref{sec:r-capability}; $\times$ in Figure~\ref{fig:dose}a) --- three of those reductions
($27$--$35\%$) being $2.4$--$3.1\times$ the replicate band of Section~\ref{sec:setup} and Credit-G's
$18\%$ only $1.6\times$.

Under Equation~\ref{eq:minimization}, deleting one inert claim at $\lambda_{\mathrm{nom}}=20$ would
raise the objective by $\rho/(|E_P|-1)$ accuracy standard deviations ($\approx 1.2$ on Credit-G,
$\approx 2.1$ on Madelon; Table~\ref{tab:dose}). This is an upper bound, since joint interventions can
absorb a deletion; a direct claim-deletion test with the pipeline held fixed is future work.

\textbf{Held-out fidelity moves independently of claim volume.} On Credit-G, uncapped held-out RCS
falls from $0.053$ at $\lambda_{\mathrm{nom}}=0$ to $0.022$ at $\lambda_{\mathrm{nom}}=5$ while claim
counts stay flat, then partly recovers to $0.044$ at $\lambda_{\mathrm{nom}}=20$. The CIs overlap and
the $\times0.42$ excursion is only marginally outside the replicate band, so it suggests fidelity moving with flat counts, not pressure-induced drift.

\subsection{Capability control: minimization is possible when instructed}
\label{sec:r-capability}

At $\lambda_{\mathrm{nom}}=5$ (Table~\ref{tab:policy}, Figure~\ref{fig:capability};
Appendix~\ref{app:tables}), the \emph{minimize} prompt reduces claim volume on four of six datasets:
Butterfly $-35\%$, Electricity $-34\%$, Madelon $-27\%$, Credit-G $-18\%$ --- three of five without
Butterfly; the other two change little (Ecommerce $-0.0\%$, Breast $-1.7\%$), so
the control is inconclusive there. Held-out RCS rises or falls
(Credit-G $\times0.60$, Butterfly $\times0.79$, Madelon $\times1.77$) within or at the edge of the
replicate band, so this arm bears on volume only. The contrast with Section~\ref{sec:r-dose} is our
central result: \textbf{shorter claim lists are reachable, but search --- Thompson sampling or UCB1 ---
did not find them without instruction.} Given feedback without claim counts and $\le 38$
proposals, this null is largely structural; the control shows only reachability.

\subsection{Probe access produces selection (F4)}
\label{sec:r-oracle}

Table~\ref{tab:oracle} and Figure~\ref{fig:oracle} (both Appendix~\ref{app:tables}) compare agents
that differ only in access to \texttt{probe\_feature\_drop}, which the probe arm calls $3.1$--$3.9$ times per
iteration against a budget of four. The selection signature stated in Section~\ref{sec:oracle} --- claim volume falls while
verified-claim rate rises --- appears on Butterfly (claims $15.5 \to 14.3$, verified rate
$0.49 \to 0.50$), Madelon ($25.6 \to 14.8$, $0.50 \to 0.51$), Credit-G ($12.4 \to 11.7$,
$0.40 \to 0.50$), and Breast ($10.5 \to 10.3$, $0.38 \to 0.41$) --- three of five without
Butterfly. The remaining two do not exhibit it:
Ecommerce's verified rate also falls ($0.54 \to 0.49$) and Electricity's claim volume rises
($8.9 \to 11.1$). Only Madelon's volume drop ($-42\%$) and Credit-G's rate rise ($+0.10$) exceed the
replicate band.

On Credit-G, held-out fidelity drops with non-overlapping
CIs ($0.034$ to $0.016$); on Ecommerce and Madelon the point estimates fall by half or more
($0.069 \to 0.024$, $0.188 \to 0.095$) but with large seed spread. All three ratios
($\times0.46$, $\times0.35$, $\times0.50$) fall below the replicate band's lower end ($\times0.61$);
reporting only the verified-claim rate would record improvement as fidelity degrades. Scoring is asymmetric: all Credit-G and two Madelon probe-access winners use
$\le 10$ features, and the estimator's small-pipeline rules (Appendix~\ref{app:rcs}) apply only to
them, so part of those declines may be scoring artifacts; Ecommerce ($40$--$76$ features)
is unaffected but has overlapping CIs.

\section{Limitations}
\label{sec:limitations}

\paragraph{Scope and power.} One verifier of our own, one generator model (Solar Pro~3), six datasets,
and pipeline-design tasks with mechanical counterfactuals make the evidence preliminary. Butterfly, our sole image task, reaches only
$\approx 0.19$ accuracy, so counts are also reported without it. Our per-dataset
trend tests detect only large effects (Section~\ref{sec:r-dose}), so we report the absence of
\emph{discovered} claim-gaming, not equivalence; pooling datasets in a mixed model is future work.

\paragraph{Generator variation and unswept settings.} Decoding parameters were fixed, but the hosted
generator's served checkpoint cannot be identified. The
replicate band of Section~\ref{sec:setup} rests on one pair of runs, so it is an order of magnitude
for fixed-seed noise, not an interval, and verdicts near its edges may change with more replicates.
Estimator constants (Appendix~\ref{app:rcs}) were not swept.

\paragraph{Normalization and untested channels.} $\rho$ can inflate on low-variance tasks without
changing selection, so we report decision impact separately (Table~\ref{tab:flip}). We did not test rephrasing that lowers the \emph{parsed} claim
count, search that observes claim count (the most direct probe of F3), and delayed or committed oracle
access; our four-call probe channel is noiseless and immediate.

\section{Conclusion}

We propose a stress-test methodology for an agent's own verifier --- the pressure ratio $\rho$, four
falsifiable probes, and a reporting standard --- that needs only an optimizing loop around a scalar
verifier; its extension to tests, task completion, or citation grounding is a hypothesis.
Across six datasets and 180 runs (plus a 90-run sweep) with one verifier and one generator,
\textbf{search did not discover claim minimization even at $\rho$ over two orders above deployment},
although instructed agents stated fewer claims --- a null that is largely structural. In summary: report $\rho$, count claims separately from scores, pair F3 robustness claims with a
capability control, and measure fidelity on verifier-unseen data.

\bibliographystyle{plainnat}
\bibliography{racle_workshop}

\appendix

\renewcommand{\topfraction}{0.95}
\renewcommand{\bottomfraction}{0.95}
\renewcommand{\textfraction}{0.05}
\renewcommand{\floatpagefraction}{0.6}
\setcounter{topnumber}{3}
\setcounter{totalnumber}{5}
\makeatletter
\setlength{\@fptop}{0pt}
\setlength{\@fpsep}{8pt}
\setlength{\@fpbot}{0pt plus 1fil}
\makeatother

\section{RCS Estimator Details}
\label{app:rcs}

The deployed estimator $\Rhat$ differs from the base estimator of Equation~\ref{eq:rcs-base} in the
ways below, and the deployed search additionally gates the verifier's weight. We state them here
because each is a degree of freedom that an evaluation of the verifier must disclose.

\paragraph{Accuracy gating of the verifier weight.} With pre-gate
$\lambda=s_d\lambda_{\mathrm{nom}}$, the weight $\lambda_t$ applied at iteration $t$
(Section~\ref{sec:pressure}) is gated on validation accuracy:
\begin{equation}
\lambda_t \;=\; \gamma_t \cdot
\begin{cases}
0 & \hat{Q}(P_t) < \tau_{\text{acc}},\ \Rhat \text{ uninformative},\\
\min(\lambda,\, c_t) & \hat{Q}(P_t) < \tau_{\text{acc}},\ \Rhat \text{ informative},\\
0.5\,\lambda & \tau_{\text{acc}} \le \hat{Q}(P_t) < 0.84,\\
1.2\,\lambda & \hat{Q}(P_t) \ge 0.84,
\end{cases}
\qquad
\gamma_t = \begin{cases} \gamma_m & t \le T_0,\\ 1 & t > T_0.\end{cases}
\label{eq:gate}
\end{equation}
Here $\gamma_m=0.78$ for tabular and $0.68$ otherwise; $\tau_{\text{acc}}=0.18/0.72/0.74/0.78$ for
image/time-series/tabular/text. The low-accuracy cap $c_t$ is $0.03/0.015$ for tabular at
$\Rhat\ge0.003/0.0015$, $0.02/0.03$ for time series at $\Rhat\ge0.01/0.003$, and $0.02$ for
image/text at $\Rhat\ge0.01$. The time-series ordering is non-monotone because the $\Rhat\ge0.01$
branch is modality-agnostic and takes precedence; both caps sit far below $\lambda_{\mathrm{nom}}$ and
apply only in the low-accuracy regime, so the objective-term effect is $\le 3\times10^{-4}$ --- below
seed-level accuracy spread. The high cutoff is $0.84$ and $T_0=\max(8,0.35T)$. Controlled sweeps
disable this gate entirely --- leaving Tables~\ref{tab:dose}--\ref{tab:dose_ucb} unaffected --- but
retain $s_d=0.98/0.98/0.88/0.82/0.88/0.78$ for image/text/time-series/Breast/
Credit-G/Madelon, so $R(P)=\hat Q(P)+s_d\lambda_{\mathrm{nom}}\Rhat(P)$.

\paragraph{Verification-aware final selection.} Beyond the objective term, the deployed system consults
the verifier once more when it selects the pipeline to report: among candidates whose validation
accuracy lies within a per-dataset tolerance ($0.004$--$0.02$, expanded to $0.035$ for Credit-G) of the best, it prefers the one
with the higher verification RCS --- ordered by its real-signal, tested-claim, and nonzero-drop counts
--- before falling back to reward and feature compactness. This tie-break is enabled in the deployed
dynamic-$\lambda$ configuration on all six datasets. It is a further channel through which the
verifier shapes the reported artifact but which $\rho$ does not register, since $\rho$
(Equation~\ref{eq:pressure}) measures only the objective term; we disclose it so that the verifier's
full influence is on record. Held-out RCS (Section~\ref{sec:setup}) remains a post-hoc diagnostic and
enters neither the objective nor this tie-break.

\paragraph{Noise threshold.} A per-claim drop at or below a modality-dependent threshold is treated
as measurement noise and contributes zero: $2.2\times10^{-4}$ for images ($1.2\times10^{-4}$ when
validation accuracy is below $0.18$), $1.8\times10^{-4}$ for text, $2.8\times10^{-4}$ for
time series, and $3.0$--$3.8\times10^{-4}$ for tabular data depending on validation accuracy. The
F1 floor rate we report is, by construction, a function of these thresholds.
Relative to one test-example accuracy unit $1/n_{\mathrm{test}}$ under the $64/16/20$ split
($n_{\mathrm{test}}\approx 0.2n$), the tabular thresholds are about $6$--$8\%$ of $1/n_{\mathrm{test}}$
on Credit-G ($n{=}1002$), order-one multiples of $1/n_{\mathrm{test}}$ on the large text and
time-series sets (Ecommerce $n{=}50427$; Electricity $n{=}45312$), and about $0.16$--$0.29$ of one
test example on Butterfly ($n{=}6499$). This places the disclosed thresholds; it is not a null
distribution for absolute held-out RCS.

\paragraph{Relative-drop boost.} When a drop exceeds a relative threshold $\tau$ of the original
validation accuracy, it is amplified by $\Delta \mapsto \Delta\,(1 + \beta\,\Delta/\hat{Q}(P))$ with
gain $\beta = 2.5$, using $\tau = 0.015$ for single-claim interventions and $\tau = 0.02$ for joint
interventions. The boost sharpens the contrast between load-bearing and inert claims; it does not
change their ordering.

\paragraph{Joint interventions.} When three or more feature-importance claims are stated (two or
more for low-accuracy image runs, where validation accuracy is below $0.10$), a subset of the named
features (one if at most ten features are selected, otherwise up to half) is removed together in a
single counterfactual and the one resulting drop is used, rather than removing each feature
separately.
This avoids crediting the same shared signal to several entangled claims, at the cost of coarser
per-claim attribution. On pipelines with at most ten selected features, every per-drop value
(joint or single-claim) is additionally capped at $0.02$.

\paragraph{Floor.} A pipeline with no parsable rationale, or with no counterfactual that could be
executed, receives a floor value of $10^{-3}$ (tabular), $5\times10^{-4}$ (text, time series), or
$2\times10^{-4}$ (image) rather than zero. The rate at which this floor is used is exactly the F1
statistic of Section~\ref{sec:results}.

\paragraph{Authoritative configuration.} The constants above, the per-dataset weight scales $s_d$, and
the selection tolerances are those set in the released \texttt{run\_experiments.py}, which is what the
runs executed. The \texttt{meta.json} written beside each unified CSV export also carries a
\texttt{racle\_budget} block, but it is a snapshot taken at export time and was not refreshed: for
Electricity it records $s_d=1.0$ and a tolerance of $0.005$ and omits the faithfulness-first flag, where
the executed configuration used $0.88$, $0.004$, and the flag enabled. Readers reconciling numbers
should use the run script, not the export metadata.

\paragraph{Held-out reporting.} Internal summaries clip held-out RCS at $0.08$ (image/text) or $0.06$
(other modalities), then at $0.035$ for pipelines with at most ten features; a legacy paper field clips
again to $[0.001,0.02]$. This paper instead uses the raw measurement recorded alongside those fields,
so comparisons across pressure levels are not truncated. ``Uncapped'' refers to these summary clips
only: the relative-drop boost and, for pipelines with at most ten features, the per-drop cap and
single-feature joint removal still apply to the raw value. Among probe-access winners, the small-pipeline rules apply to all
three on Credit-G ($8$ features) and two of three on Madelon ($10$), but to
no control winner ($11$--$17$ and $11$--$26$ features), which is why Section~\ref{sec:r-oracle}
qualifies those declines; comparing pre-cap, pre-boost drops across arms is future work.

\section{Datasets}
\label{app:data}

Table~\ref{tab:datasets} lists the six tasks; all are public. Splits and preprocessing follow
Section~\ref{sec:setup}.

\paragraph{Unified export artifact.} Every task is read from a unified CSV export, and in four of them
that export repeated the header line twice as data. Credit-G, Breast Cancer, Madelon, and Ecommerce
were therefore trained on two spurious rows each, which is what the $n$ in
Table~\ref{tab:datasets} records; their canonical sizes are $1000$, $286$, $2000$ (Madelon's training split), and $50{,}425$.
The spurious rows form a third label with two examples --- under $0.7\%$ of even the smallest affected
task --- and are identical across every arm, seed, and pressure level, so they cannot produce a
between-arm difference. Butterfly and Electricity are unaffected.

\paragraph{Meaning of $n$.} For five tasks $n$ is the row count of the single ingested file, split
$64/16/20$ afterwards. Butterfly is the exception: its source distributes a labelled training manifest
($6499$ rows, the value in Table~\ref{tab:datasets}) and a separate $2786$-row manifest without
labels. Only the labelled manifest is ingested, so Butterfly's $n$ counts labelled rows
available rather than all rows shipped with the dataset.

\paragraph{Tabular.} Credit-G is the Statlog German Credit dataset~\citep{hofmann1994german}
(OpenML \texttt{credit-g}, id~31; German bank applicants, $20$ attributes, good/bad-risk labels). Breast
Cancer is the UCI Ljubljana breast-cancer recurrence dataset~\citep{zwitter1988breast}. Madelon is
the artificial NIPS~2003 feature-selection challenge dataset~\citep{guyon2004result} ($500$ features,
of which $480$ are non-informative noise probes), which we use as a high-dimensional sparsity stress
test.

\paragraph{Time series.} Electricity is the ELEC2 dataset~\citep{harries1999splice} from the
Australian New South Wales electricity market ($45{,}312$ half-hourly records; binary up/down price
movement). We construct lag-1/2/3 features on five series (23 features after expansion) and use a
time-ordered rather than random split.

\paragraph{Text.} Ecommerce Text is the E-commerce text dataset of \citet{gautam2019ecommerce}, a
product-description corpus in four categories (Household, Books, Electronics, Clothing \&
Accessories), represented by unigram/bigram TF-IDF features.

\paragraph{Image.} Butterfly Image is the public Kaggle Butterfly Image Classification
dataset~\citep{phucthai_butterfly};
each image is represented by simple grayscale-intensity statistics and histograms (26 dimensions,
with a filename-hash fallback when the raw image is unavailable), not by pretrained-CNN embeddings. The
classifier reaches only $\approx 0.19$ test accuracy on this 75-class task --- well above the
$1/75 \approx 0.013$ chance rate, but low in absolute terms, since the representation is deliberately
weak. Because it is by a wide margin the lowest-accuracy task in the suite, we report all dataset
counts both with and without Butterfly as a robustness check.

\section{Full Result Tables}
\label{app:tables}

\paragraph{Identical-configuration replicate.} Table~\ref{tab:replicate} isolates the run-to-run
variation quoted in Section~\ref{sec:setup}. Two arms of our design coincide exactly: the
$\lambda_{\mathrm{nom}}=5$ level of the dose--response sweep and the \emph{default} arm of the
capability control both run with static $\lambda_{\mathrm{nom}}=5$, seeds $42/43/44$, the default
proposer, and no policy prompt. They were nonetheless executed as two independent invocations, so their
gap measures how much the generator alone moves the reported quantities at a fixed seed --- a component
the across-seed bootstrap cannot see, because each of its resampled runs draws a fresh generator
sample. We therefore treat claim-volume differences under $11.4\%$, held-out RCS ratios within
$[\times0.61, \times2.71]$, verified-rate differences under $0.06$, and test-accuracy differences under $0.048$ as indistinguishable from
re-execution noise, and say so wherever an effect falls inside that band.

\begin{table}[t]
\centering
\caption{Identical-configuration replicate. Run~A is the $\lambda_{\mathrm{nom}}=5$ arm of the dose--response sweep (Table~\ref{tab:dose}); Run~B is the \emph{default} arm of the capability control (Table~\ref{tab:policy}). Both fix static $\lambda_{\mathrm{nom}}=5$ and seeds $42/43/44$ and apply no policy prompt, so they differ only in being invoked twice. Three-seed means; held-out RCS is uncapped.}
\label{tab:replicate}
\small
\setlength{\tabcolsep}{4.5pt}
\renewcommand{\arraystretch}{1.15}
\begin{tabular}{lcccccc}
\toprule
& \multicolumn{3}{c}{Claims/iter} & \multicolumn{2}{c}{Held-out RCS} & Test acc.\ \\
\cmidrule(lr){2-4}\cmidrule(lr){5-6}\cmidrule(lr){7-7}
Dataset & A & B & $\Delta$ & A $\to$ B & ratio & $|\Delta|$ \\
\midrule
 Credit-G & 10.6 & 11.5 & $+7.7\%$ & 0.0222 $\to$ 0.0235 & $\times$1.06 & 0.010 \\
 Breast & 10.2 & 9.9 & $-2.7\%$ & 0.0099 $\to$ 0.0061 & $\times$0.61 & 0.011 \\
 Madelon & 20.3 & 18.9 & $-6.7\%$ & 0.0783 $\to$ 0.0766 & $\times$0.98 & 0.022 \\
 Ecommerce & 9.5 & 10.5 & $+11.4\%$ & 0.1493 $\to$ 0.1691 & $\times$1.13 & 0.048 \\
 Butterfly & 13.9 & 14.5 & $+4.1\%$ & 0.0554 $\to$ 0.0792 & $\times$1.43 & 0.017 \\
 Electricity & 9.1 & 9.6 & $+5.1\%$ & 0.0049 $\to$ 0.0134 & $\times$2.71 & 0.002 \\
\bottomrule
\end{tabular}
\end{table}

\paragraph{Selection flip rate.} Table~\ref{tab:flip} reports decision impact, the quantity $\rho$ does
not measure. For every run we re-select the reported pipeline from the iterations that run actually
executed, once under $\hat Q(P_t)+\lambda_t\Rhat(P_t)$ and once under $\hat Q(P_t)$ alone, and record
whether the choice changes and how much validation accuracy the changed choice gives up. Two
qualifications matter. It is a post-hoc re-selection, not a re-run: deleting the verifier term would
also change which pipelines the search proposed, so the flip rate characterizes the observed candidate
pool rather than a counterfactual trajectory. And it compares objective terms only, so the
final-selection tie-break of Appendix~\ref{app:rcs} is excluded. Read with those limits, the
$\lambda_{\mathrm{nom}}=0$ row is a null check (no term, no flips), the deployed row shows influence
confined to accuracy ties, and the sweep rows show the elevated $\rho$ values translating into
decisions taken at real accuracy cost.

\begin{table}[t]
\centering
\caption{Selection flip rate: how often the verifier term decides which pipeline a run reports. For every run we re-select over the iterations it actually executed, comparing $\arg\max_t\,[\hat Q(P_t)+\lambda_t\Rhat(P_t)]$ with $\arg\max_t\,\hat Q(P_t)$. \emph{Acc.\ conceded} is the validation accuracy the flipped choice gives up, so $0.000$ means the term broke a tie. Post-hoc re-selection, not a re-run: removing the term would also change the trajectory.}
\label{tab:flip}
\small
\setlength{\tabcolsep}{5pt}
\renewcommand{\arraystretch}{1.15}
\begin{tabular}{lccc}
\toprule
Arm & Runs & Flip rate & Acc.\ conceded (mean/max) \\
\midrule
 deployed (dynamic $\lambda$) & 18 & 5/18 (28\%) & 0.000 / 0.000 \\
 $\lambda_{\mathrm{nom}}=0$ & 18 & 0/18 (0\%) & 0.000 / 0.000 \\
 $\lambda_{\mathrm{nom}}=0.12$ & 18 & 6/18 (33\%) & 0.003 / 0.009 \\
 $\lambda_{\mathrm{nom}}=1$ & 18 & 10/18 (56\%) & 0.011 / 0.044 \\
 $\lambda_{\mathrm{nom}}=5$ & 18 & 14/18 (78\%) & 0.034 / 0.209 \\
 $\lambda_{\mathrm{nom}}=20$ & 18 & 13/18 (72\%) & 0.036 / 0.192 \\
\addlinespace[3pt]
 P2 control (no probe) & 18 & 4/18 (22\%) & 0.002 / 0.003 \\
 P2 treatment (probe) & 18 & 3/18 (17\%) & 0.000 / 0.000 \\
\bottomrule
\end{tabular}
\end{table}

\begin{figure}[!ht]
\centering
\includegraphics[width=0.92\linewidth]{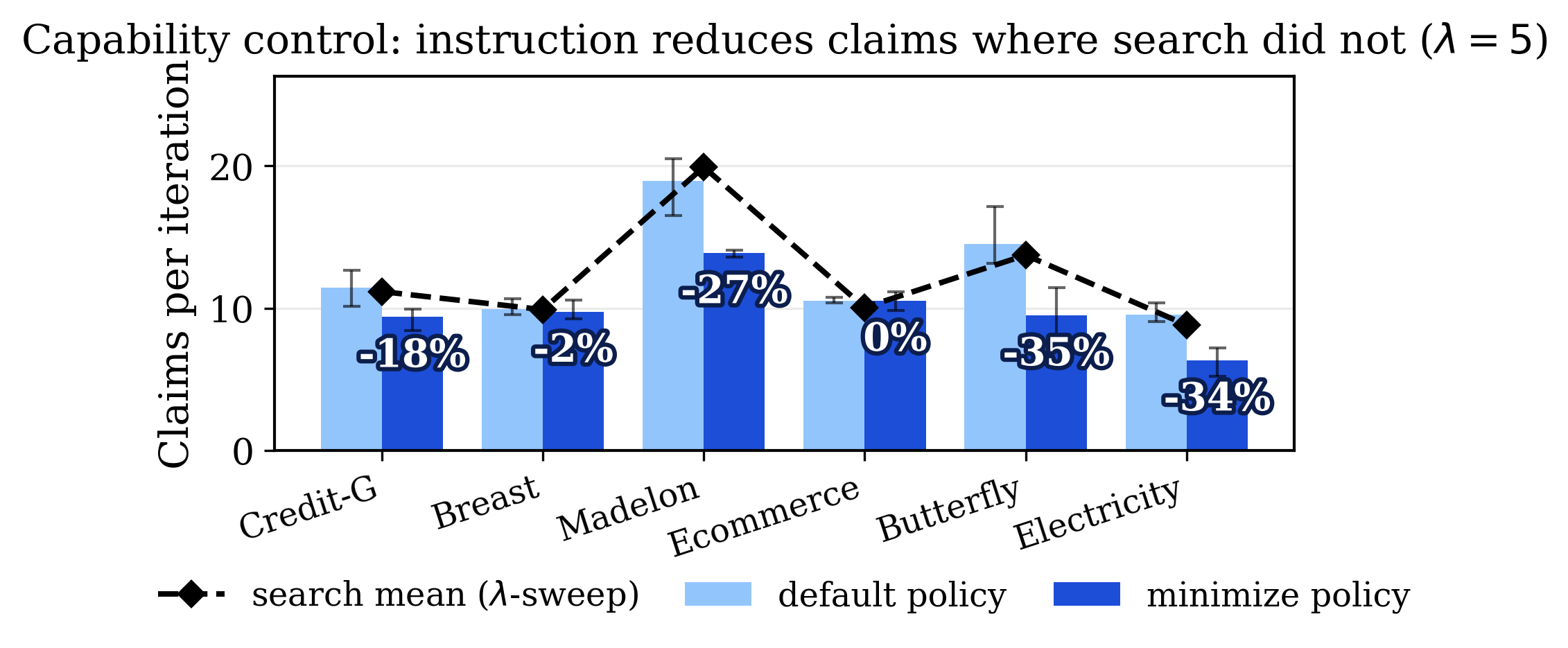}
\caption{Capability control at $\lambda_{\mathrm{nom}}=5$: claim volume under the default vs.\ minimize policy
(bars, bootstrap 95\% CIs). The dashed line is the mean claim volume reached by the nominal-$\lambda$ sweep
under search; instruction drives volume below it on the four datasets where search never did. Percent
labels give the minimize reduction relative to default.}
\label{fig:capability}
\end{figure}

\begin{figure}[!ht]
\centering
\includegraphics[width=0.70\linewidth]{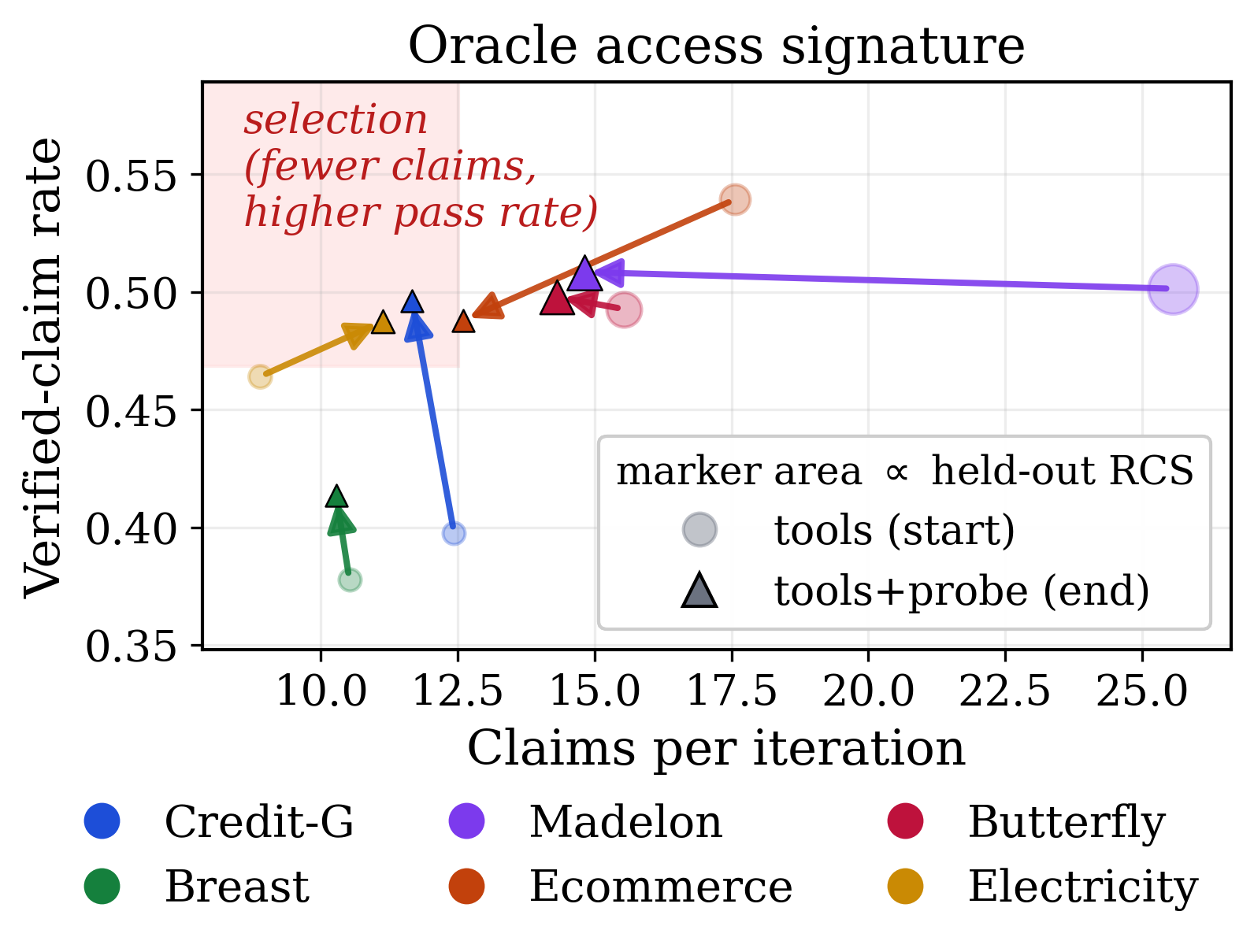}
\caption{Oracle access signature (three seeds averaged). Each arrow runs from the \texttt{tools} arm
(circle) to the \texttt{tools+probe} arm (triangle); marker area is proportional to uncapped held-out
RCS. The shaded upper-left quadrant is the selection regime --- fewer claims at a higher verified
rate. Arrows point into it on four datasets; held-out fidelity (marker area) drops with
non-overlapping CIs only on Credit-G and is suggestive but noisy across seeds elsewhere.}
\label{fig:oracle}
\end{figure}

Tables~\ref{tab:policy}--\ref{tab:dose} give the full per-dataset capability-control
(Figure~\ref{fig:capability}), probe-access (Figure~\ref{fig:oracle}), and dose--response
(Figure~\ref{fig:dose}) results (three-seed means, bootstrap $95\%$ CIs, uncapped held-out RCS;
percent reductions quoted in the text are computed from unrounded means);
Table~\ref{tab:dose_ucb} repeats the dose--response under UCB1 with a doubled base cap of 30
iterations (realized $18$--$38$; Section~\ref{sec:r-dose}).

\begin{table}[t]
\centering
\caption{Capability control at nominal $\lambda_{\mathrm{nom}}=5$: default/minimize claim policies,
three seeds.}
\label{tab:policy}
\small
\setlength{\tabcolsep}{6pt}
\renewcommand{\arraystretch}{1.15}
\begin{tabular}{lcccc}
\toprule
Dataset & Policy & Claims/iter & Verified rate & Held-out RCS \\
\midrule
 \multirow{2}{*}{Credit-G} & default & 11.5 & 0.52 & 0.023 [0.008, 0.033] \\
  & minimize & 9.4 & 0.45 & 0.014 [0.009, 0.022] \\
\addlinespace[3pt]
 \multirow{2}{*}{Breast} & default & 9.9 & 0.36 & 0.006 [0.004, 0.010] \\
  & minimize & 9.8 & 0.41 & 0.010 [0.001, 0.020] \\
\addlinespace[3pt]
 \multirow{2}{*}{Madelon} & default & 18.9 & 0.57 & 0.077 [0.006, 0.123] \\
  & minimize & 13.9 & 0.60 & 0.136 [0.101, 0.162] \\
\addlinespace[3pt]
 \multirow{2}{*}{Ecommerce} & default & 10.5 & 0.62 & 0.169 [0.088, 0.253] \\
  & minimize & 10.5 & 0.62 & 0.120 [0.067, 0.203] \\
\addlinespace[3pt]
 \multirow{2}{*}{Butterfly} & default & 14.5 & 0.49 & 0.079 [0.064, 0.107] \\
  & minimize & 9.5 & 0.35 & 0.063 [0.049, 0.079] \\
\addlinespace[3pt]
 \multirow{2}{*}{Electricity} & default & 9.6 & 0.64 & 0.013 [0.011, 0.015] \\
  & minimize & 6.3 & 0.67 & 0.018 [0.010, 0.033] \\
\bottomrule
\end{tabular}
\end{table}

\begin{table*}[t]
\centering
\caption{Probe access to the verifier, three seeds. Held-out RCS is uncapped.}
\label{tab:oracle}
\footnotesize
\setlength{\tabcolsep}{3pt}
\renewcommand{\arraystretch}{1.12}
\begin{tabular}{llccccc}
\toprule
Dataset & Arm & Probes/iter & Claims/iter & Verified rate & Test acc.\ & Held-out RCS \\
\midrule
 \multirow{2}{*}{Credit-G} & tools & 0.0 & 12.4 & 0.40 & 0.718 [0.682, 0.756] & 0.034 [0.025, 0.051] \\
  & tools+probe & 3.1 & 11.7 & 0.50 & 0.725 [0.701, 0.741] & 0.016 [0.010, 0.020] \\
\addlinespace[3pt]
 \multirow{2}{*}{Breast} & tools & 0.0 & 10.5 & 0.38 & 0.690 [0.672, 0.724] & 0.008 [0.004, 0.011] \\
  & tools+probe & 3.6 & 10.3 & 0.41 & 0.701 [0.655, 0.724] & 0.009 [0.004, 0.014] \\
\addlinespace[3pt]
 \multirow{2}{*}{Madelon} & tools & 0.0 & 25.6 & 0.50 & 0.792 [0.771, 0.810] & 0.188 [0.142, 0.250] \\
  & tools+probe & 3.4 & 14.8 & 0.51 & 0.787 [0.758, 0.808] & 0.095 [0.005, 0.274] \\
\addlinespace[3pt]
 \multirow{2}{*}{Ecommerce} & tools & 0.0 & 17.6 & 0.54 & 0.657 [0.592, 0.780] & 0.069 [0.042, 0.101] \\
  & tools+probe & 3.8 & 12.6 & 0.49 & 0.678 [0.591, 0.803] & 0.024 [0.001, 0.067] \\
\addlinespace[3pt]
 \multirow{2}{*}{Butterfly} & tools & 0.0 & 15.5 & 0.49 & 0.196 [0.192, 0.201] & 0.093 [0.084, 0.100] \\
  & tools+probe & 3.9 & 14.3 & 0.50 & 0.195 [0.190, 0.199] & 0.090 [0.078, 0.111] \\
\addlinespace[3pt]
 \multirow{2}{*}{Electricity} & tools & 0.0 & 8.9 & 0.46 & 0.703 [0.692, 0.709] & 0.040 [0.010, 0.055] \\
  & tools+probe & 3.6 & 11.1 & 0.49 & 0.729 [0.709, 0.770] & 0.041 [0.014, 0.055] \\
\bottomrule
\end{tabular}
\end{table*}

\begin{table*}[t]
\centering
\caption{Dose--response to nominal static verifier weight, three seeds; dataset scale $s_d$ remains
active. Held-out RCS is uncapped and $\rho$ measures pressure.}
\label{tab:dose}
\footnotesize
\setlength{\tabcolsep}{3.5pt}
\renewcommand{\arraystretch}{1.12}
\begin{tabular}{llcccccc}
\toprule
Dataset & $\lambda_{\rm nom}$ & $\rho$ & Claims/iter & Verified rate & Chars & Held-out RCS \\
\midrule
 \multirow{5}{*}{Credit-G} & 0 & 0.00 & 10.8 & 0.51 & 282 & 0.053 [0.038, 0.066] \\
  & 0.12 & 0.07 & 11.3 & 0.56 & 333 & 0.047 [0.034, 0.068] \\
  & 1 & 1.05 & 12.2 & 0.52 & 297 & 0.019 [0.012, 0.033] \\
  & 5 & 4.42 & 10.6 & 0.48 & 320 & 0.022 [0.005, 0.051] \\
  & 20 & 11.58 & 11.0 & 0.53 & 285 & 0.044 [0.030, 0.070] \\
\addlinespace[3pt]
 \multirow{5}{*}{Breast} & 0 & 0.00 & 9.9 & 0.42 & 698 & 0.009 [0.005, 0.012] \\
  & 0.12 & 0.04 & 9.8 & 0.42 & 715 & 0.010 [0.001, 0.015] \\
  & 1 & 0.30 & 9.7 & 0.32 & 663 & 0.007 [0.005, 0.010] \\
  & 5 & 1.19 & 10.2 & 0.37 & 665 & 0.010 [0.007, 0.015] \\
  & 20 & 4.96 & 10.0 & 0.41 & 664 & 0.009 [0.005, 0.015] \\
\addlinespace[3pt]
 \multirow{5}{*}{Madelon} & 0 & 0.00 & 20.3 & 0.57 & 447 & 0.080 [0.005, 0.124] \\
  & 0.12 & 0.28 & 20.3 & 0.58 & 431 & 0.079 [0.006, 0.150] \\
  & 1 & 1.78 & 17.7 & 0.58 & 448 & 0.115 [0.095, 0.148] \\
  & 5 & 9.00 & 20.3 & 0.63 & 421 & 0.078 [0.008, 0.142] \\
  & 20 & 41.66 & 21.0 & 0.54 & 507 & 0.117 [0.103, 0.125] \\
\addlinespace[3pt]
 \multirow{5}{*}{Ecommerce} & 0 & 0.00 & 10.9 & 0.64 & 337 & 0.069 [0.067, 0.072] \\
  & 0.12 & 0.06 & 9.8 & 0.68 & 323 & 0.072 [0.066, 0.078] \\
  & 1 & 0.52 & 9.6 & 0.58 & 285 & 0.164 [0.007, 0.269] \\
  & 5 & 2.95 & 9.5 & 0.63 & 293 & 0.149 [0.058, 0.196] \\
  & 20 & 10.35 & 10.6 & 0.61 & 320 & 0.138 [0.080, 0.247] \\
\addlinespace[3pt]
 \multirow{5}{*}{Butterfly} & 0 & 0.00 & 13.5 & 0.44 & 274 & 0.091 [0.047, 0.121] \\
  & 0.12 & 0.05 & 12.2 & 0.47 & 267 & 0.057 [0.046, 0.070] \\
  & 1 & 0.46 & 14.9 & 0.45 & 276 & 0.066 [0.063, 0.069] \\
  & 5 & 7.10 & 13.9 & 0.45 & 298 & 0.055 [0.049, 0.059] \\
  & 20 & 11.63 & 14.2 & 0.42 & 288 & 0.061 [0.044, 0.073] \\
\addlinespace[3pt]
 \multirow{5}{*}{Electricity} & 0 & 0.00 & 9.1 & 0.68 & 269 & 0.009 [0.004, 0.019] \\
  & 0.12 & 0.12 & 8.6 & 0.59 & 288 & 0.018 [0.015, 0.022] \\
  & 1 & 1.59 & 8.6 & 0.66 & 263 & 0.016 [0.015, 0.019] \\
  & 5 & 12.51 & 9.1 & 0.64 & 315 & 0.005 [0.001, 0.012] \\
  & 20 & 63.66 & 8.7 & 0.55 & 263 & 0.008 [0.001, 0.014] \\
\bottomrule
\end{tabular}
\end{table*}

\begin{table}[!ht]
\centering
\caption{Dose--response under deterministic UCB1 with the base cap doubled to 30 before applying the
dataset-specific iteration-budget multiplier
(typically 34--38 iterations; two $\lambda_{\rm nom}=0$ runs used 18 and 26). The nominal
static-weight sweep matches Table~\ref{tab:dose}; held-out RCS is uncapped and $\rho$ measures pressure.
Claim-count curves are nonmonotonic in $\lambda_{\rm nom}$.}
\label{tab:dose_ucb}
\footnotesize
\setlength{\tabcolsep}{3.5pt}
\renewcommand{\arraystretch}{1.0}
\begin{tabular}{llcccccc}
\toprule
Dataset & $\lambda_{\rm nom}$ & $\rho$ & Claims/iter & Verified rate & Chars & Held-out RCS \\
\midrule
 \multirow{5}{*}{Credit-G} & 0 & 0.00 & 10.6 & 0.52 & 290 & 0.045 [0.039, 0.051] \\
  & 0.12 & 0.12 & 10.7 & 0.53 & 275 & 0.028 [0.015, 0.043] \\
  & 1 & 0.55 & 11.2 & 0.54 & 300 & 0.021 [0.005, 0.039] \\
  & 5 & 6.46 & 10.1 & 0.46 & 281 & 0.029 [0.024, 0.034] \\
  & 20 & 9.12 & 10.7 & 0.55 & 278 & 0.035 [0.025, 0.047] \\
\addlinespace[1.5pt]
 \multirow{5}{*}{Breast} & 0 & 0.00 & 10.0 & 0.38 & 634 & 0.010 [0.001, 0.015] \\
  & 0.12 & 0.03 & 9.7 & 0.36 & 673 & 0.006 [0.001, 0.015] \\
  & 1 & 0.28 & 9.8 & 0.43 & 658 & 0.014 [0.011, 0.015] \\
  & 5 & 1.44 & 9.8 & 0.38 & 653 & 0.009 [0.005, 0.015] \\
  & 20 & 5.51 & 9.8 & 0.43 & 689 & 0.011 [0.005, 0.015] \\
\addlinespace[1.5pt]
 \multirow{5}{*}{Madelon} & 0 & 0.00 & 20.3 & 0.58 & 442 & 0.080 [0.006, 0.158] \\
  & 0.12 & 0.28 & 17.3 & 0.61 & 454 & 0.118 [0.004, 0.187] \\
  & 1 & 1.83 & 18.1 & 0.63 & 441 & 0.179 [0.148, 0.217] \\
  & 5 & 14.61 & 20.0 & 0.56 & 428 & 0.142 [0.004, 0.249] \\
  & 20 & 38.69 & 18.1 & 0.55 & 423 & 0.151 [0.130, 0.185] \\
\addlinespace[1.5pt]
 \multirow{5}{*}{Ecommerce} & 0 & 0.00 & 10.9 & 0.62 & 345 & 0.141 [0.064, 0.294] \\
  & 0.12 & 0.05 & 10.0 & 0.58 & 305 & 0.054 [0.002, 0.097] \\
  & 1 & 0.59 & 9.9 & 0.64 & 323 & 0.155 [0.016, 0.260] \\
  & 5 & 3.02 & 11.3 & 0.63 & 350 & 0.190 [0.145, 0.223] \\
  & 20 & 16.41 & 10.6 & 0.62 & 324 & 0.088 [0.066, 0.103] \\
\addlinespace[1.5pt]
 \multirow{5}{*}{Butterfly} & 0 & 0.00 & 13.4 & 0.44 & 274 & 0.095 [0.078, 0.121] \\
  & 0.12 & 0.08 & 15.9 & 0.42 & 281 & 0.068 [0.064, 0.073] \\
  & 1 & 0.88 & 13.9 & 0.39 & 279 & 0.091 [0.046, 0.142] \\
  & 5 & 3.51 & 15.4 & 0.49 & 310 & 0.096 [0.066, 0.142] \\
  & 20 & 12.24 & 15.4 & 0.45 & 288 & 0.094 [0.057, 0.142] \\
\addlinespace[1.5pt]
 \multirow{5}{*}{Electricity} & 0 & 0.00 & 8.2 & 0.70 & 272 & 0.014 [0.007, 0.020] \\
  & 0.12 & 0.21 & 8.6 & 0.67 & 287 & 0.026 [0.018, 0.030] \\
  & 1 & 2.48 & 8.4 & 0.54 & 278 & 0.019 [0.011, 0.025] \\
  & 5 & 9.34 & 8.1 & 0.69 & 292 & 0.012 [0.003, 0.018] \\
  & 20 & 38.43 & 8.4 & 0.65 & 320 & 0.012 [0.009, 0.014] \\
\bottomrule
\end{tabular}
\end{table}

\end{document}